\documentclass[10pt,twocolumn,letterpaper]{article}

\usepackage{cvpr}
\usepackage{times}
\usepackage{soul}
\usepackage{url}
\usepackage[hidelinks]{hyperref}
\usepackage[utf8]{inputenc}
\usepackage[small]{caption}
\usepackage{graphicx}
\usepackage{amsmath}
\usepackage{booktabs}
\usepackage{algorithm}
\usepackage{algorithmic}
\usepackage{epsfig}
\usepackage{amssymb}
\usepackage{color}
\usepackage{multirow}
\usepackage{subcaption}
\usepackage{tabularx} 
\usepackage{flushend}
\usepackage[table]{xcolor}

\cvprfinalcopy 

\def\cvprPaperID{216} 

\ifcvprfinal\fi
\begin{document}

\title{Multi-Prototype Networks for Unconstrained Set-based Face Recognition}

\author{\normalsize{Jian~Zhao$^{1,2,3}$\thanks{Jian Zhao is the corresponding author. Homepage: \url{https://zhaoj9014.github.io}. This work was done during Jian Zhao served as a short-term ``Texpert" research scientist at Tencent FiT DeepSea AI Lab, Shenzhen, China.}, Jianshu~Li$^{1}$, Xiaoguang~Tu$^{1,4}$, Fang~Zhao$^{5}$, Yuan~Xin$^{3}$, Junliang~Xing$^{6}$, Hengzhu~Liu$^{2}$, Shuicheng~Yan$^{1,7}$, Jiashi~Feng$^{1}$} \\
	\small{$^{1}$National University of Singapore, $^{2}$National University of Defense Technology} \\
		\small{$^{3}$Tencent FiT DeepSea AI Lab, $^{4}$University of Electronic Science and Technology of China} \\ \small{$^{5}$Inception Institute of Artificial Intelligence, $^{6}$Institute of Automation, Chinese Academy of Sciences, $^{7}$Qihoo 360 AI Institute} \\
	{\small  \{zhaojian90, jianshu\}@u.nus.edu, xguangtu@outlook.com, zhaofang0627@gmail.com, macxin@tencent.com} \\ {\small jlxing@nlpr.ia.ac.cn, hengzhuliu@nudt.edu.cn, \{eleyans, elefjia\}@nus.edu.sg}}

\maketitle

\begin{abstract}
	In this paper, we study the challenging unconstrained set-based face recognition problem where each subject face is instantiated by a set of media (images and videos) instead of a single image. Naively aggregating information from all the media within a set would suffer from the large intra-set variance caused by heterogeneous factors (\emph{e.g.}, varying media modalities, poses and illuminations) and fail to learn discriminative face representations. A novel \textbf{M}ulti-\textbf{P}rototype \textbf{Net}work (MPNet) model is thus proposed to learn multiple prototype face representations adaptively from the media sets. Each learned prototype is representative for the subject face under certain condition in terms of pose, illumination and media modality. Instead of handcrafting the set partition for prototype learning, MPNet introduces a \textbf{D}ense \textbf{S}ub\textbf{G}raph (DSG) learning sub-net that implicitly untangles inconsistent media and learns a number of representative prototypes. Qualitative and quantitative experiments clearly demonstrate superiority of the proposed model over state-of-the-arts.
\end{abstract}


\section{Introduction}

\begin{figure}[t]
	\begin{center}
		\includegraphics[width=1\linewidth]{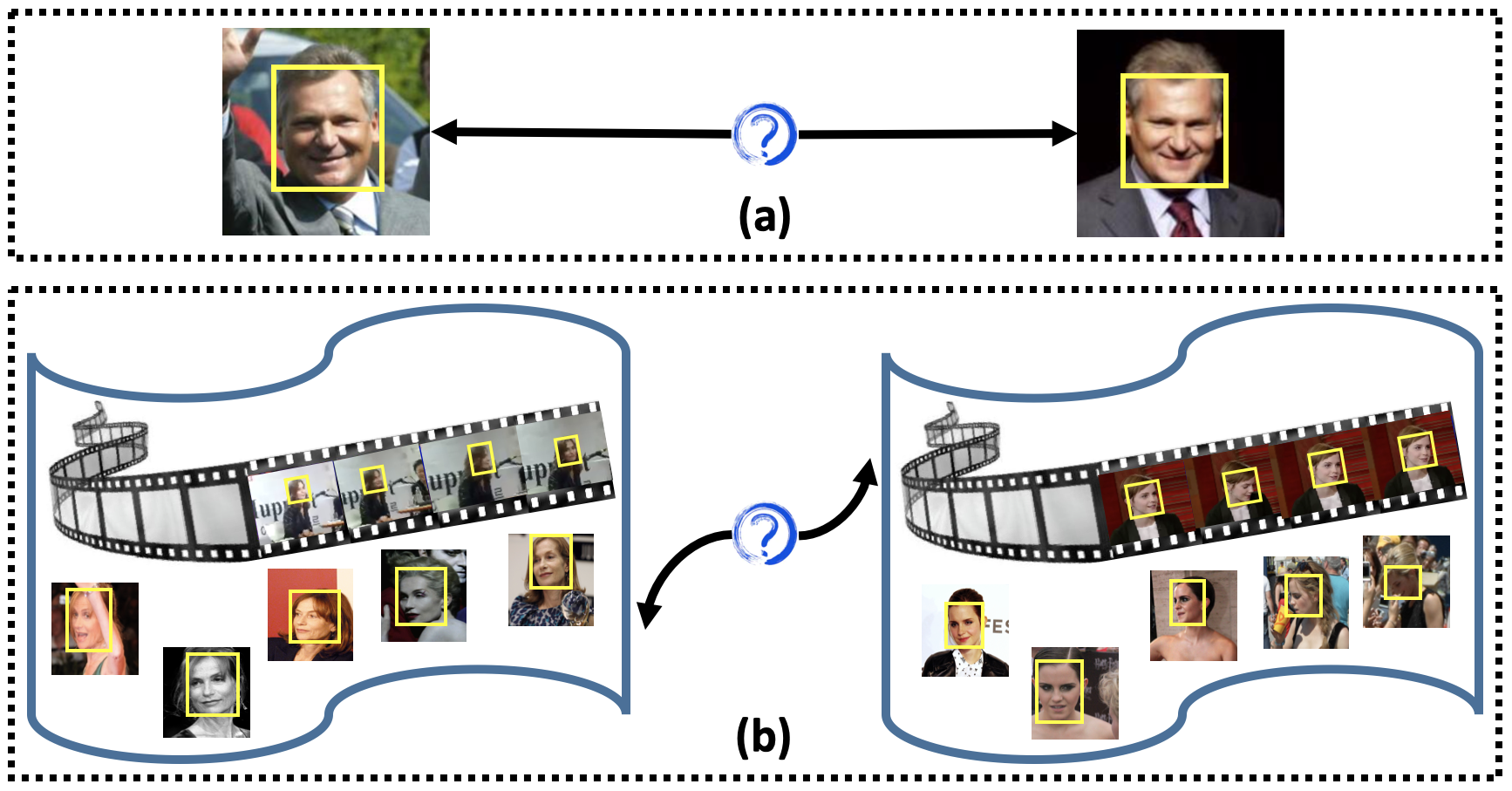}
	\end{center}
	\vspace{-4mm}
	\caption{\small Difference between (a) face recognition over a single image per subject and (b) unconstrained set-based face recognition. For unconstrained set-based face recognition, each subject is represented by a set of mixed images and videos captured under unconstrained conditions. Each set contains large variations in face pose, expression, illumination and occlusion issues. Existing single-medium based recognition approaches cannot successfully address this problem. Best viewed in color.}
	\label{fig: Figure1}
\end{figure}

Recent advances of deep learning approaches have remarkably boosted the performance of face recognition. Some approaches claim to have achieved~\cite{Taigman:Deepface,chen2017robust,li2016robust,zhao2017dual,zhao2018towards} or even surpassed~\cite{Schroff:Facenet,wang2018cosface,ijcai2018-165} human performance on several benchmarks. However, those approaches only recognize faces over a single image or video sequence. Such scenarios deviate from  the reality. In practical face recognition systems (and arguably human cortex for face recognition), each subject face to recognize is often enrolled with \emph{a set} of images and videos captured under varying conditions and acquisition methods. Intuitively such rich information can benefit face recognition performance, which however has not been effectively exploited in existing approaches~\cite{Hao:ITRC-SARI,Lu:extensionFACENET,zhao2017dual}. 

In this paper, we consider the challenging task\textemdash \emph{unconstrained set-based face recognition} firstly introduced in~\cite{Klare:ijba}\textemdash  that is more consistent with real-world scenarios. Unconstrained set-based face recognition defines the minimal facial representation unit as a set of images and videos instead of a single medium. Set-based face recognition requires solving a more difficult set-to-set matching problem, where both the probe and gallery are sets of face media. This  task raises the necessity to build \emph{subject-specific} face models for each subject individually, instead of relying on a single multi-class recognition model as before. An illustration on the difference between traditional face recognition over a single input image and the targeted face recognition over a set of unconstrained images/videos is given in Fig.~\ref{fig: Figure1}. The most significant challenge in the unconstrained set-based face recognition task is how to learn good representations for the media set, even in presence of  \emph{large intra-set variance} of real-world subject faces caused by varying conditions in illumination, sensor, compression, \textit{etc.}, and subject attributes such as facial pose, expression and occlusion. Solving this problem needs to address these distracting factors effectively and learn set-level discriminative face representation. 

Recently, several set-based face recognition methods have been proposed~\cite{Chellappa:Recognition,Chen:Unconstrainedfaceverification,Chowdhury:BCNNs,Hassner:Poolingfaces}. They generally adopt the following two strategies to obtain set-level face representation. One is to learn a set of image-level face representations from each face medium in the set individually~\cite{Chowdhury:BCNNs,Masi:Poseaware}, and use all the information for following face recognition. Such a strategy is obviously computationally expensive as it needs to perform exhaustive pairwise matching and is fragile to outlier medium captured under unusual conditions. The other strategy is to aggregate face representations within the set through simple average or max pooling and generate single representation for each set~\cite{Chen:Unconstrainedfaceverification,sankaranarayanan2016triplet}. However, this  obviously suffers from information loss and insufficient exploitation of the image/video set.

To overcome the limitations of existing methods, we propose a novel \textbf{M}ulti-\textbf{P}rototype \textbf{Net}work (MPNet) model. To learn better set-level representations, 
MPNet introduces a \textbf{D}ense \textbf{S}ub\textbf{G}raph (DSG) learning sub-net to implicitly factorize each face media set of a particular subject into a number of disentangled sub-sets, instead of handcrafting the set partition using some intuitive features. Each dense subgraph discovers a sub-set (representing a \emph{prototype}) of face media that are with small intra-set variance but discriminative from other subject faces. MPNet learns to enhance the compactness of the prototypes as well as their coverage of large variance for a single subject face, through which heterogeneous attributes within each face media set are sufficiently considered and flexibly untangled. This significantly helps improve the unconstrained set-based face recognition performance by providing multiple comprehensive and succinct face representations, reducing the impact of media inconsistency. Compared with existing set-based face recognition methods~\cite{Chellappa:Recognition,Chen:Unconstrainedfaceverification,Chowdhury:BCNNs,Hassner:Poolingfaces}, MPNet effectively addresses the large variance challenge and offers more discriminative and flexible face representations with lower computational complexity. Also, superior to naive average or max pooling of face features, MPNet effectively preserves the necessary information through the DSG learning for set-based face recognition. The main contributions of this work can be summarized  as follows:
\begin{itemize}
	\setlength\itemsep{0em}
	\item We propose a novel and effective multi-prototype discriminative learning architecture MPNet. To our best knowledge, MPNet is the first end-to-end trainable model that adaptively learns multiple prototype face representations from sets of media. It is effective at  addressing the   large intra-set variance issue that is critical to set-based face recognition. 
	\item MPNet introduces a \textbf{D}ense \textbf{S}ub\textbf{G}raph (DSG) learning sub-net that  automatically factorizes each face media set into a number of disentangled prototypes representing consistent face media with sufficient discriminativeness. Through the DSG sub-net, MPNet is capable of untangling inconsistent media and dealing with faces captured under challenging conditions robustly.
	\item  DSG provides a general loss that encourages compactness around multiple discovered centers with strong discrimination. It offers a new and systematic approach for large variance object recognition in the real world.
\end{itemize}
Based on the above technical contributions, we have presented a high-performance model for unconstraint set-based face recognition. It achieves currently best results on IJB-A~\cite{Klare:ijba}, YTF~\cite{wolf:ytf} and IJB-C~\cite{maze2018iarpa} benchmark datasets with significant improvement over state-of-the-arts.

\section{Related Work}

\begin{figure*}[t]
	\begin{center}
		\includegraphics[width=0.8\linewidth]{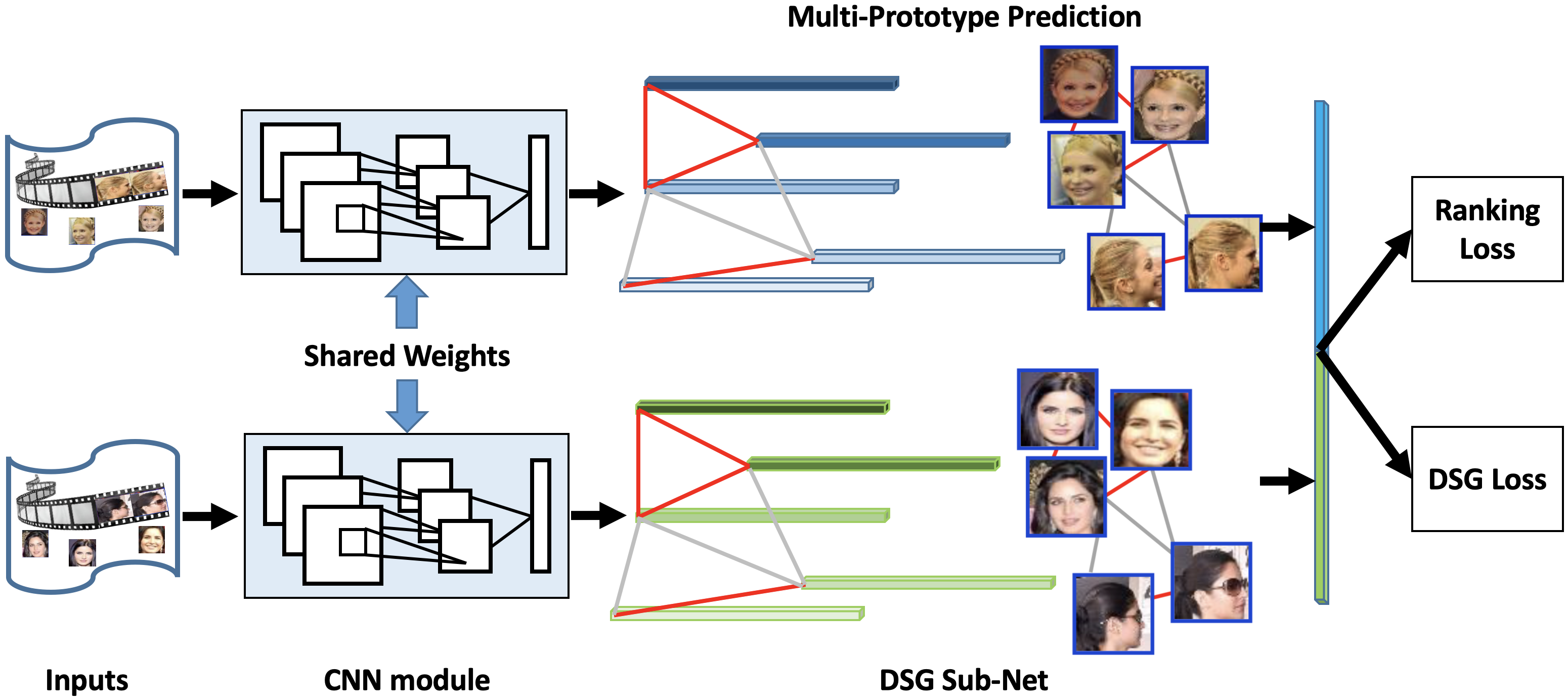}
	\end{center}
	\vspace{-4mm}
	\caption{\small The proposed \textbf{M}ulti-\textbf{P}rototype \textbf{Net}work (MPNet) for unconstrained set-based face recognition. MPNet takes a pair of face media sets as input and outputs a matching result, \emph{i.e.}, same person or not. It adopts a modern deep siamese CNN architecture for deep set-based facial representation learning, and introduces a new DSG sub-net to learn discriminative  prototypes for each set. MPNet is end-to-end trainable through the   ranking loss and   auxiliary DSG loss.  Best viewed in color.}
	\label{fig: Figure3}
\end{figure*}

Recent top performing approaches for face recognition often rely on deep CNNs with advanced architectures. For instance, the VGGface model~\cite{Parkhi15,cao2018vggface2}, as an application of the VGG architecture~\cite{simonyan:vgg}, provides state-of-the-art performance. The DeepFace model~\cite{Taigman:Deepface,taigman:web} also uses a deep CNN coupled with 3D alignment. FaceNet~\cite{Schroff:Facenet} utilizes the inception deep CNN architecture for unconstrained face recognition. DeepID2+~\cite{sun:deepid2} and DeepID3~\cite{sun:deepid3} extend the FaceNet model by including joint Bayesian metric learning and multi-task learning, yielding better unconstrained face recognition performance. SphereFace~\cite{liu2017sphereface}, CosFace~\cite{wang2018cosface}, AM-Softmax~\cite{wang2018additive} and ArcFace~\cite{deng2018arcface} exploit margin-based representation learning to achieve small intra-class distance and large inter-class distance. Those methods enhance their overall performance via carefully designed architectures, which are however not tailored for unconstrained set-based face recognition.

With the introduction of IJB-A benchmark~\cite{Klare:ijba} by NIST in $2015$, the problem of unconstrained set-based face recognition attracts increasing attention. Recent solutions to this problem are also based on deep architectures, which are leading approaches on LFW~\cite{huang:lfw} and YTF~\cite{wolf:ytf}. Among them, B-CNN~\cite{Chowdhury:BCNNs} applies a new \textbf{B}ilinear \textbf{CNN} (B-CNN) architecture for face identification. Pooling Faces~\cite{Hassner:Poolingfaces} aligns faces in 3D and partitions them according to facial and imaging properties. PAMs~\cite{Masi:Poseaware} handles pose variability by learning \textbf{P}ose-\textbf{A}ware \textbf{M}odel\textbf{s} (PAMs) for frontal, half-profile and full-profile poses to perform face recognition in the wild. Those methods often employ separate processing steps without considering the modality variance within one set of face media and underlying multiple prototype structures. Therefore, much useful information may loss, leading to inferior performance.

Our proposed MPNet shares a similar idea as subcategory-aware object classification~\cite{Dong:Subcategory} that considers intra-class variance in building object classifiers, and \cite{das2018sample,das2018unsupervised} that considers to construct a graph-matching metric for domain adaption. Our method differs from them in following aspects:~1) the ``prototype" is not pre-defined in MPNet; 2) learning DSG within each face media set implicitly discovers consistent faces sharing similar conditions, through which heterogeneous factors are flexibly untangled; 3) MPNet is based on deep learning and can be end-to-end trainable. It is also interesting to investigate the application of our MPNet architecture in generic object recognition tasks.

\section{Multi-Prototype Networks}

Fig.~\ref{fig: Figure3} visualizes the architecture of the MPNet, which takes a pair of face media sets as input and outputs a matching result for the unconstrained set-based face recognition. It adopts a modern deep siamese CNN architecture for set-based facial representation learning, and introduces a new DSG sub-net for learning the multi-prototype that models various representative faces under different conditions for the input. MPNet is end-to-end trainable by minimizing the ranking loss and a new DSG loss. We now present each component in detail.

\subsection{Set-based Facial Representation Learning}

Different from face recognition over a single image, the task of set-based face recognition aims to accept or reject the claimed identity of a subject represented by a face media set containing both images and videos. Performance is assessed using two measures: percentage of false accepts and that of false rejects. A good model should optimize both metrics simultaneously. MPNet is designed to nonlinearly map the raw sets of faces to multiple prototypes in a low dimensional space such that the distance between these prototypes is small if the sets belong to the same subject, and large otherwise. The similarity metric learning is achieved by training MPNet with two identical CNN branches that share weights. MPNet handles inputs in a pair-wise, set-to-set way so that it explicitly organizes the face media in a way favorable to set-based face recognition.

\begin{figure}[t]
	\begin{center}
		\includegraphics[width=0.9\linewidth]{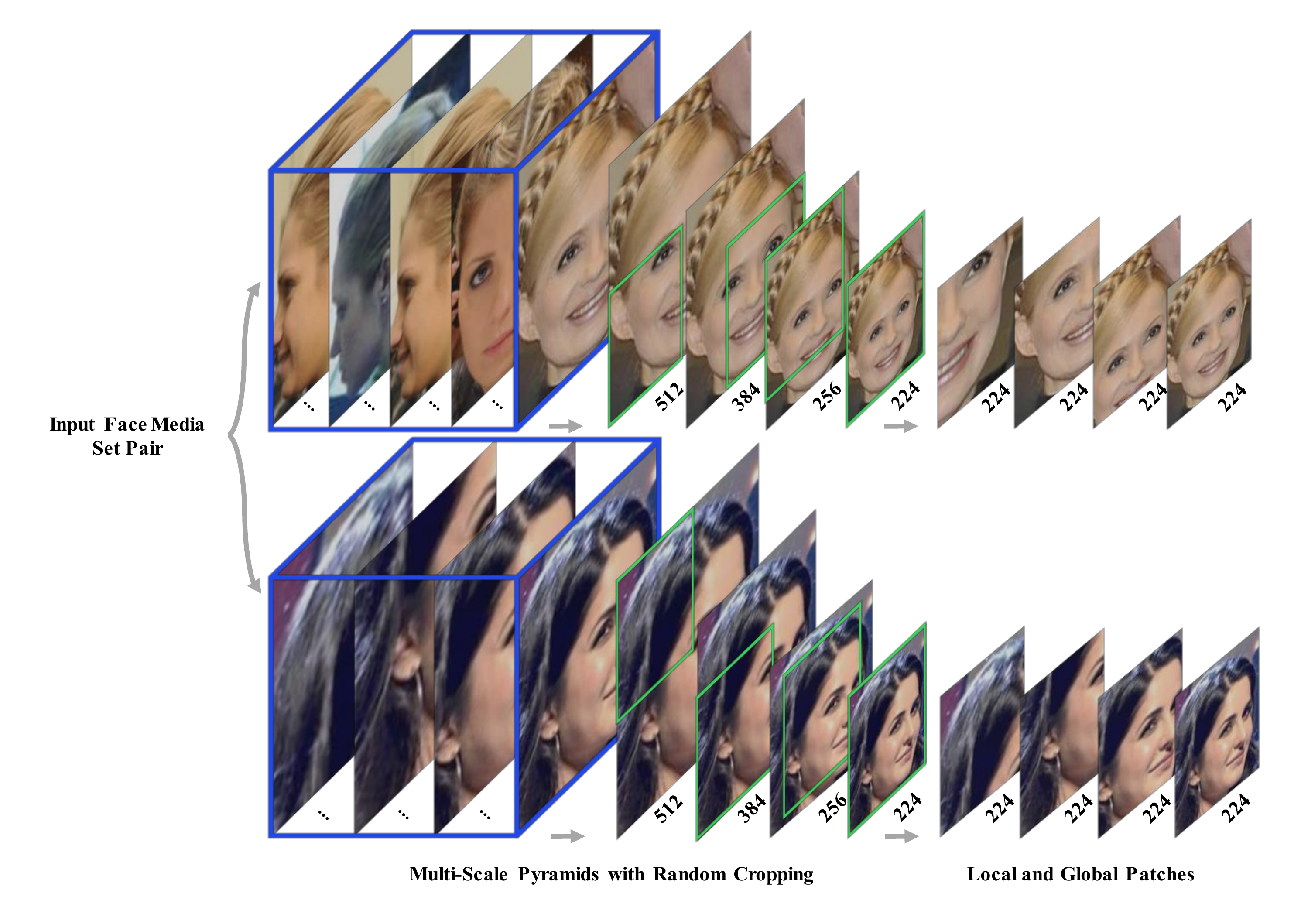}
	\end{center}
	\vspace{-4mm}
	\caption{\small Illustration of multi-scale pyramid construction and random cropping strategy for the MPNet. For each medium within a face media set, a multi-scale pyramid is constructed by resizing the medium to four different scales. Local and global patches are then randomly cropped from each multi-scale pyramid with a fixed size. Best viewed in color.}
	\label{fig: Figure311}
\end{figure}

MPNet learns face representations at multi-scale for gaining strengthened robustness to scale variance in real-world faces. Specifically, for each medium within a face media set, a multi-scale pyramid is constructed by resizing the image or video frame to four different scales. To handle the error of face detection, MPNet performs random cropping to collect local and global patches from each scale of the multi-scale pyramid with a fixed size, as illustrated in Fig.~\ref{fig: Figure311}. To handle the imbalance of realistic face data (\emph{e.g.}, some subjects are enrolled with scarce media from limited images while some  with redundant media from reduplicative video frames), the data distribution within each set is adjusted by resampling. In particular, the set containing scarce media (\emph{i.e.}, less than a pre-defined parameter ${R}$ that is set empirically) is augmented by duplicating and flipping images, which is intuitively beneficial with the support from more relevant information. The large set with redundant media (\emph{i.e.}, more than ${R}$) is subsampled to the size of ${R}$. The resulting input streams to MPNet are tuples of face media set pairs and the associated ground truth annotations $\{(X^{p1},X^{p2},y^p)\}$, where $X^{p1}$ and $X^{p2}$ denote the two sets of the $p$-th pair, and $y^p$ denotes the binary pair-wise label.

The proposed MPNet adopts a siamese CNN architecture in which two branches share weights for pairwise set-based facial representation learning. Each branch is initiated with VGGface~\cite{Parkhi15}, including 13 convolutional layers, 5 pooling layers and 2 fully-connected layers. We make the following careful architectural design for each branch to ensure that the learned deep facial representations are more suitable for multi-prototype representation learning.~1) For activation functions, instead of using ReLU~\cite{nair:relu} to suppress all the negative responses, we adopt the PReLU~\cite{he:prelu} to allow negative responses. PReLU improves model fitting with nearly zero extra computational cost and little overfitting risk, benefiting  convergence of MPNet. 2) We adopt two local normalization layers after the $2^{nd}$ convolutional layer and the $4^{th}$ convolutional layer, respectively. The local normalization tends to uniformize the mean and variance of a feature map around a local neighborhood. This is especially useful for correcting non-uniform illumination or shading artifacts.
3) We adopt an average operator for the last pooling layer and a max operator for the previous $4$ pooling layers to generate compact and discriminative deep set-based facial representations. Note that our approach is not restricted to the CNN module used, and can also be generalized to other state-of-the-art architectures for performance boosting.

The learned deep facial representation for each face media set is denoted as
$\{f_1,f_2,\cdots,f_R\}$. Here recall $R$ is the specified size of the face media set after distribution balance.

\subsection{Multi-Prototype Discriminative Learning}

Throughout this paper, a prototype is defined as a collection of similar face media that are representative for a subject face under certain conditions. Face media forming the same prototype have small variance and one can safely extract representation by pooling approaches without worrying about information loss.  

To address the critical large variance issue in set-based face recognition, we propose the multi-prototype discriminative learning. With this component, each face media set is implicitly factorized into a certain number of prototypes. Multi-prototype learning encourages the output facial representations to be compact w.r.t. certain prototypes and meanwhile discriminative for different subjects. Thus, MPNet is capable of modelling the prototype-level interactions effectively while addressing the large variance and false matching caused by untypical faces. MPNet dynamically offers an optimal trade-off between facial information preserving and computation cost. It does not require exhaustive matching across all of the possible pairs from two sets for accurately recognizing faces in the wild. It learns multiple prototypes through a dense subgraph learning as detailed below.

To discover the underlying multiple prototypes of each face media set instead of handcrafting the set partition, we propose a novel DSG learning approach. DSG formulates the similarity of face media within a set through a graph and discovers the prototype by mining the dense subgraphs. Each subgraph has high internal similarity and small similarity to the outside media. Compared with clustering-based data partition, DSG is advantageous in terms of flexiblity and robustness to outliers. Each subgraph provides a prototype for the input subject faces. We then perform face recognition at the prototype level, which is concise and also sufficiently informative. 

Given a latent affinity graph characterizing the relations among entities (face media here), denoted as $\mathcal{G}$=$(\mathcal{V},\mathcal{E})$, each dense subgraph refers to a prototype of the vertices ($\mathcal{V}$) with larger internal connection ($\mathcal{E}$) affinity than other candidates. In this work, learning DSG within each face media set implicitly discovers consistent faces sharing similar conditions such as age, pose, expression, illumination and media modality, through which heterogeneous factors are flexibly untangled.

Formally, suppose the graph $\mathcal{G}$ is associated with an affinity  $A$, and each element of $A$ encodes the similarity between two face media: $a_{ij}=\mathrm{aff}(f_i,f_j)$. Let $K$ be the number of prototypes (or equivalently, number of dense subgraphs) and $Z=[z_1,\ldots,z_K] \in \mathbb{R}^{n\times K}$ be the partition indicator: $z_{ik}=1$ indicates the $i$-th medium is  in the $k$-th prototype. The DSG aims to find the representative subgraph via optimizing $Z$ through 

\begin{equation}
\small
\label{eqn:DSG}
\max_{Z}\mathrm{tr}(Z^\top AZ), \text{ s.t. }, z_{ik} \in \{0,1\}, Z\mathbf{1}=\mathbf{1}, 
\end{equation}
where $\mathbf{1}$ is an all-1 vector. Here the $2^{nd}$ constraint guarantees that every medium will be allocated to only one prototype. The allocation after learning would maximize the intra-prototype media similarity. This is significantly different from k-means clustering where the centers are not necessary to learn and similarity is not defined based on the distance to the center.

This problem is not easy to solve. We therefore carefully design the DSG layers that form the DSG sub-net to optimize it end-to-end. This sub-net consists of two layers, which takes the  set-based facial representations $f_i$'s as input and outputs the reconstructed discriminative features. 

The $1^{st}$ layer makes the prototype prediction. Given the input deep facial representation $f_i$, this layer outputs its indicator $z_i \in \{0,1\}$ by dynamically projecting each input latent affinity graph to $K$ prototypes, $z_i = \sigma(W^\top f_i)$, where $\sigma$ is the sigmoid activation function to rectify inputs to $[0, 1]$ and $W$ is the DSG predictor parameter. Given the predicted $z_i$ and input $f_i$, the $2^{nd}$ layer computes the DSG loss (see Sec.~\ref{sec3.3}) to ensure the reconstructed representation $\hat{f_i}$ form reasonably compact and discriminative prototypes. $\hat{f_i}$ is obtained by element-wisely multiplying the output of the $2^{nd}$ layer of the DSG sub-net with the predicted multi-prototype indicator $Z$ from its $1^{st}$ layer. More details are given in Sec.~\ref{Sec4}.

\begin{table*}
	\newcommand{\tabincell}[2]{\begin{tabular}{@{}#1@{}}#2\end{tabular}}
	\tiny
	\begin{center}
		\begin{tabular}{p{4.1cm}<{\centering}|p{1.7cm}<{\centering}|p{1.7cm}<{\centering}|p{1.7cm}<{\centering}||p{1.7cm}<{\centering}|p{1.7cm}<{\centering}|p{1.7cm}<{\centering}}
			\hline
			\multirow{2}{*}{\tabincell{c}{\textbf{Method}}} &\multicolumn{3}{c||}{\textbf{Verification}} &\multicolumn{3}{c}{\textbf{Identification}} \\
			\cline{2-7}
			& \tabincell{c}{\textbf{TAR@FAR=0.10}} & \tabincell{c}{\textbf{TAR@FAR=0.01}} & \tabincell{c}{\textbf{TAR@FAR=0.001}} & \tabincell{c}{\textbf{FNIR@FPIR=0.10}} & \tabincell{c}{\textbf{FNIR@FPIR=0.01}} & \textbf{Rank1} \\
			\hline\hline
			OpenBR~\cite{Klare:ijba} & 0.433$\pm$0.006 & 0.236$\pm$0.009 & 0.104$\pm$0.014 & 0.851$\pm$0.028 & 0.934$\pm$0.017 & 0.246$\pm$0.011 \\
			GOTS~\cite{Klare:ijba} & 0.627$\pm$0.012 & 0.406$\pm$0.014 & 0.198$\pm$0.008 & 0.765$\pm$0.033 & 0.953$\pm$0.024 & 0.433$\pm$0.021 \\
			BCNNs~\cite{Chowdhury:BCNNs} & - & - & - & 0.659$\pm$0.032 & 0.857$\pm$0.024 & 0.588$\pm$0.020 \\
			Pooling faces~\cite{Hassner:Poolingfaces} & 0.631 & 0.309 & - & - & - & 0.846 \\
			LSFS~\cite{wang2015face} & 0.895$\pm$0.013 & 0.733$\pm$0.034 & 0.514$\pm$0.060 & 0.387$\pm$0.032 & 0.617$\pm$0.063 & 0.820$\pm$0.024 \\
			Deep Milti-pose~\cite{AbdAlmageed:multi} & 0.991 & 0.787 & - & 0.250 & 0.48 & 0.846 \\
			DCNN$_{\mathrm{manual}}$+metric~\cite{chen2015end} & 0.947$\pm$0.011 & 0.787$\pm$0.043 & - & - & - & 0.852$\pm$0.018 \\
			Triplet Similarity~\cite{sankaranarayanan2016triplet} & 0.945$\pm$0.002 & 0.790$\pm$0.030 & 0.590$\pm$0.050 & 0.246$\pm$0.014 & 0.444$\pm$0.065 & 0.880$\pm$0.015 \\
			PAMs~\cite{Masi:Poseaware} & - & 0.826$\pm$0.018 & 0.652$\pm$0.037 & - & - & 0.840$\pm$0.012 \\
			DCNN$_{\mathrm{fusion}}$~\cite{Chen:Unconstrainedfaceverification} & 0.967$\pm$0.009 & 0.838$\pm$0.042 & - & 0.210$\pm$0.033 & 0.423$\pm$0.094 & 0.903$\pm$0.012 \\
			Masi \emph{et al.}~\cite{Masi:facerecognition} & - & 0.886 & 0.725 & - & - & 0.906 \\
			Triplet Embedding~\cite{sankaranarayanan2016triplet} & 0.964$\pm$0.005 & 0.900$\pm$0.010 & 0.813$\pm$0.002 & 0.137$\pm$0.014 & 0.247$\pm$0.030 & 0.932$\pm$0.010 \\
			All-In-One~\cite{ranjan2016all} & 0.976$\pm$0.004 & 0.922$\pm$0.010 & 0.823$\pm$0.020 & 0.113$\pm$0.014 & 0.208$\pm$0.020 & 0.947$\pm$0.008 \\
			Template Adaptation~\cite{Crosswhite:Templateadaptation} & 0.979$\pm$0.004 & 0.939$\pm$0.013 & 0.836$\pm$0.027 & 0.118$\pm$0.016 & 0.226$\pm$0.049 & 0.928$\pm$0.001 \\
			NAN~\cite{yang2017neural} & 0.979$\pm$0.004 & 0.941$\pm$0.008 & 0.881$\pm$0.011 & 0.083$\pm$0.009 & 0.183$\pm$0.041 & 0.958$\pm$0.005 \\
			DA-GAN~\cite{zhao2017dual} & 0.991$\pm$0.003 & 0.976$\pm$0.007 & 0.930$\pm$0.005 & 0.051$\pm$0.009 & 0.110$\pm$0.039 & 0.971$\pm$0.007 \\
			$\ell_2$-softmax~\cite{ranjan2017l2} & 0.984$\pm$0.002 & 0.970$\pm$0.004 & 0.943$\pm$0.005 & 0.044$\pm$0.006 & 0.085$\pm$0.041 & 0.973$\pm$0.005 \\
			3D-PIM~\cite{ijcai2018-165} & 0.996$\pm$0.001 & 0.989$\pm$0.002 & 0.977$\pm$0.004 & 0.016$\pm$0.005 & 0.064$\pm$0.045 & 0.990$\pm$0.002 \\ 
			\hline
			\rowcolor{gray!25}
			baseline & 0.968$\pm$0.009 & 0.871$\pm$0.014 & 0.735$\pm$0.031 & 0.188$\pm$0.011 & 0.372$\pm$0.045 & 0.907$\pm$0.010 \\
			\rowcolor{gray!25}
			w/o DSG & 0.971$\pm$0.006 & 0.887$\pm$0.012 & 0.743$\pm$0.027 & 0.182$\pm$0.010 & 0.367$\pm$0.041 & 0.912$\pm$0.008 \\
			\rowcolor{gray!25}
			MPNet$_{\mathrm{K}=3}$ & 0.971$\pm$0.006 & 0.863$\pm$0.019 & 0.734$\pm$0.033 & 0.189$\pm$0.013 & 0.386$\pm$0.043 & 0.909$\pm$0.007 \\
			\rowcolor{gray!25}
			MPNet$_{\mathrm{K}=10}$ & 0.971$\pm$0.007 & 0.880$\pm$0.015 & 0.740$\pm$0.026 & 0.179$\pm$0.009 & 0.361$\pm$0.044 & 0.913$\pm$0.009 \\
			\rowcolor{gray!25}
			MPNet$_{\mathrm{K}=200}$ & 0.979$\pm$0.004 & 0.924$\pm$0.013 & 0.764$\pm$0.022 & 0.171$\pm$0.012 & 0.350$\pm$0.046 & 0.923$\pm$0.008 \\
			\rowcolor{gray!25}
			MPNet$_{\mathrm{K}=500}$ & 0.980$\pm$0.005 & 0.919$\pm$0.013 & 0.779$\pm$0.021 & 0.169$\pm$0.009 & 0.337$\pm$0.042 & 0.932$\pm$0.008 \\
			\rowcolor{gray!25}
			MPNet$_{\mathrm{K}=1000}$ & 0.975$\pm$0.008 & 0.909$\pm$0.017 & 0.757$\pm$0.025 & 0.164$\pm$0.011 & 0.359$\pm$0.040 & 0.926$\pm$0.010 \\
			\hline
			\textbf{MPNet (Ours)} & \textbf{0.997$\pm$0.002} & \textbf{0.991$\pm$0.003} & \textbf{0.984$\pm$0.005} & \textbf{0.011$\pm$0.005} & \textbf{0.059$\pm$0.040} & \textbf{0.994$\pm$0.003} \\
			\hline
		\end{tabular}
	\end{center}
	\vspace{-4mm}
	\caption{\small Face recognition performance comparison on IJB-A. The results are averaged over 10 testing splits. ``-" means the result is not reported. Standard deviation is not available for some methods.}
	\label{tab: Table1}
\end{table*} 

\subsection{Optimization}
\label{sec3.3}

We optimize MPNet by minimizing the following two loss functions in a conjugate way.

\textbf{Ranking loss:} a ranking loss is designed in MPNet to enforce the distance to shrink for genuine set pairs and be large for imposter set pairs, so that MPNet explicitly maps input patterns into the target spaces to approximate the semantic distance in the input space. 

We first $\ell_2$-normalize the outputs from the DSG sub-net, so that all the set-based facial representations are within the same range for loss computation. Then, we use Euclidean distance to measure the fine-grained pairwise dissimilarity between $\hat{f_i}$ and $\hat{f_j}$:

\begin{equation}
\small
\label{eqn:distance}
d_{ij} = \|\hat{f_i} - \hat{f_j}\|^2_2,
\end{equation} 
where $i,j\in\{1,...,{R}\}$.

We further ensemble the ${R}^2$ distances into one energy-based matching result for each coarse-level set pair:

\begin{equation}
\small
\label{eqn:energy}
E = \dfrac{\sum_{i,j}^{R}d_{ij}\times  \exp(\beta d_{ij})}{\sum_{i,j}^{R}\exp(\beta d_{ij})},
\end{equation}
where $\beta$ is a bandwidth parameter. 

Our final ranking loss function is formulated as   

\begin{equation}\label{eqn:ranking}
\small
\mathcal{L}_{\mathrm{Ranking}}(\hat{f_i})  \triangleq \min_{E} \left\{(1-y^p)E+y^p\max(0,\tau-E)\right\},
\end{equation} 
where $\tau$ is a margin, such that $E_G+\tau<E_I$, $E_G$ is the distance for genuine pair, $E_I$ is the distance for imposter pair, and $y^p$ is the binary pairwise label, with $0$ for genuine pair ($\mathcal{L}_{\mathrm{Ranking}}=E_G$ in Eq.~\eqref{eqn:ranking}) and $1$ for imposter pair ($\mathcal{L}_{\mathrm{Ranking}}=\max(0, \tau - E_I)$ in Eq.~\eqref{eqn:ranking}).

\textbf{Dense SubGraph loss:} We propose to learn dense prototypes through solving the problem defined in Eq.~\eqref{eqn:DSG}. Expanding the objective in Eq.~\eqref{eqn:DSG} gives

\begin{equation}\label{eqn:tr}
\small
\mathrm{tr}(Z^\top AZ) = \sum_{i,j=1}^n \sum_{k=1}^K  z_{ik}a_{ij}z_{jk}.
\end{equation}

Since $z_{ik}, z_{jk} \in \{0,1\}$, we have $\sum_{k=1}^K  z_{ik}a_{ij}z_{jk} = a_{ij}$ only if $z_{ik}=z_{jk}=1$ for some $k$, \emph{i.e.}, the face media $i$ and $j$ are divided into the same prototype. Maximizing the trace in Eq.~\eqref{eqn:DSG} is to find the partition of samples (indicated by $z$) to form subgraphs such that the samples associated with the same subgraph have the largest total affinity (\emph{i.e.}, density) $\sum_{z_{ik}\neq 0, z_{jk} \neq 0} a_{ij}$. In practice, maximizing $\mathrm{tr}(Z^\top AZ)$ would encourage contributions of the representations $\hat{f_i}$ belonging to the same prototype to be close to each other and each resulted cluster to be far away from others, \emph{i.e.}, they form multiple dense subgraphs. 

In Eq.~\eqref{eqn:tr}, each element of the  affinity $A$ encodes similarity between two corresponding media, \emph{i.e.}

\begin{equation}
\small
a_{ij} \triangleq \exp\{\dfrac{-d_{ij}}{\delta^2}\}.
\end{equation}

Equivalently, the DSG learning can be achieved through the following minimization problem:

\begin{equation}
\small
\min_{Z} \mathrm{tr}(Z^\top DZ), \text{ s.t. } z_{ij} \in \{0,1\}, Z\mathbf{1}=\mathbf{1},
\end{equation}
where $D$ is the Euclidean distance matrix computed in Eqn.~\eqref{eqn:distance}. 

Then we define the following DSG loss function to optimize the learned deep set-based facial representations:

\begin{equation}
\small
\mathcal{L}_{\mathrm{DSG}}(\hat{f_i}) \triangleq \left\{\min_{Z} \mathrm{tr}(Z^\top D Z), \text{ s.t. } z_{ij} \in \{0,1\}, Z\mathbf{1}=\mathbf{1} \right \}.
\end{equation}

Thus, minimizing the DSG loss would encourage contributions of the representations $\hat{f_i}$'s belonging to the same prototype to be close to each other. If one visualizes the learned representations in the high-dimensional space, the learned representations of one face media set form several \emph{compact} clusters and each cluster may be far away from others. In this way, a face media set with large variance is distributed to several clusters implicitly. Each cluster has a small variance. We also conduct experiments for illustration in Sec.~\ref{sec4.1}.

To simplify the above optimization, we propose to relax the constraint of $z_{ij} \in \{0,1\}$  to $0 \leq  z_{ij} \leq  1$ by a sigmoid activation function. Thus, the DSG loss is re-defined as

\begin{equation}
\small
\label{eqn:dsg_loss}
\mathcal{L}_{\mathrm{DSG}}(\hat{f_i}) \triangleq \left\{\min_{Z} \mathrm{tr}(Z^\top DZ), \text{ s.t. } 0 \leq  z_{ij} \leq  1, Z\mathbf{1}=\mathbf{1} \right \}.
\end{equation}

We adopt the joint supervision of ranking and DSG losses to train MPNet for multi-prototype discriminative learning:

\begin{equation}
\small
\mathcal{L} = \mathcal{L}_{\mathrm{Ranking}} + \lambda \mathcal{L}_{\mathrm{DSG}},
\end{equation}
where $\lambda$ is a weighting parameter among the two losses.

Clearly, MPNet is end-to-end trainable and can be optimized with BP and SGD algorithm. We summarize the learning algorithm of MPNet in Algorithm~\ref{alg: Algorithm1}. 

\begin{algorithm*}[t] 
	\renewcommand{\algorithmicrequire}{ \textbf{Input:}} 
	\renewcommand{\algorithmicensure}{ \textbf{Output:}} 
	\caption{Multi-prototype learning algorithm}  
	\begin{algorithmic}[t]  
		\REQUIRE Training data $X_p=\{(X^{p1},X^{p2}, y^p)\}$. Initialized parameters $\theta$, $W$ in the CNN module and DSG sub-net, respectively. Hyperparameters $R, K, \beta, \tau, \lambda$ and learning rate $\mu^t$. The number of iteration $t\leftarrow 0$;  
		\ENSURE The parameters $\theta$ and $W$.\\ 
		\WHILE {not converge}  
		\STATE $t\leftarrow t+1$;  
		\STATE Compute the joint loss by $\mathcal{L}^t = \mathcal{L}_{Ranking}^t + \lambda \mathcal{L}_{DSG}^t$;  
		\STATE Compute the backpropagation error for each p by $\frac{\partial \mathcal{L}^t}{\partial X_p^t}=\frac{\partial \mathcal{L}_{Ranking}^t}{\partial X_p^t}+\lambda \cdot \frac{\partial \mathcal{L}_{DSG}^t}{\partial X_p^t}$;
		\STATE Update the parameters $\theta$ by $\theta^{t+1}=\theta^t - \mu^t\Sigma^m_p{\frac{\partial \mathcal{L}^t}{\partial X_p^t}\cdot \frac{\partial X_p^t}{\partial \theta^t}}$;
		\STATE Update the parameters $W$ by $W^{t+1}=W^t - \mu^t\Sigma^m_p{\frac{\partial \mathcal{L}^t}{\partial X_p^t}\cdot \frac{\partial X_p^t}{\partial W^t}}$;
		\ENDWHILE     
	\end{algorithmic}  
	\label{alg: Algorithm1}
\end{algorithm*} 

\rowcolors{2}{gray!25}{white}
\begin{table}
	\newcommand{\tabincell}[2]{\begin{tabular}{@{}#1@{}}#2\end{tabular}}
	\tiny
	\begin{center}
		\begin{tabular}{p{4.2cm}<{\centering}|p{3.2cm}<{\centering}}
			\hline
			\tabincell{c}{\textbf{Method}} & \tabincell{c}{\textbf{Acc}} \\
			\hline\hline
			DeepID~\cite{sun2014deep} & 0.932 \\
			DeepFace~\cite{Taigman:Deepface} & 0.914 \\
			Center loss~\cite{wen2016discriminative} & 0.949 \\
			SphereFace~\cite{liu2017sphereface} & 0.950 \\
			FaceNet~\cite{Schroff:Facenet} & 0.951 \\
			VGGface~\cite{Parkhi15} & 0.973 \\
			CosFace~\cite{wang2018cosface} & 0.976 \\
			ArcFace~\cite{deng2018arcface}  & 0.980 \\
			\hline
			\textbf{MPNet (Ours)} & \textbf{0.991} \\
			\hline
		\end{tabular}
	\end{center}
	\vspace{-4mm}
	\caption{\small Face recognition performance comparison on YTF.}
	\label{tab: Table2}
\end{table}

\section{Experiments}
\label{Sec4}

We evaluate MPNet qualitatively and quantitatively under various settings for unconstrained set-based face recognition on IJB-A~\cite{Klare:ijba}, YTF~\cite{wolf:ytf} and IJB-C~\cite{maze2018iarpa}.

\paragraph{Implementation Details} We initialize the CNN module of MPNet for deep set-based facial representation learning with the VGGface model~\cite{Parkhi15}, and fine-tune it on the target dataset. For each medium with the provided face bounding box, we first crop the facial RoI accordingly and then resize it to multiple $r\times r\times 3$ sizes to build the multi-scale pyramids, where $r{=}224, 256, 384$ and $512$. The  size of inputs to MPNet is fixed as $224\times 224\times 3$ by randomly cropping local and global patches of compatible size from  images/video frames. No 2D or 3D face alignment is used. The threshold $R$ for balancing input data distribution is set as 128 for trading-off recognition accuracy and computation cost. The weights of the $1^{st}$ layer (implemented with a 1D convolution layer with sigmoid activation function) of the DSG sub-net are initialized by normal distribution with an std 0.001. The number of total prototypes $K$ is set as 500. We also conduct experiments to illustrate how the $K$ influences the overall performance in Sec. 4.2. The bandwidth parameter $\beta$ in Eq.~\eqref{eqn:energy} is set to 10, the margin $\tau$ of the ranking loss is fixed as 0.8, and the trade-off parameter $\lambda$ is set as 0.01 by 5-fold cross-validation. Different values of $\lambda$ lead to different deep feature distributions. With proper $\lambda$, the discriminative power of deep features can be significantly enhanced. $\lambda=0.01$ is large enough for balancing the scales of two loss terms as the sub-graph loss calculates summations over more pairs. The proposed network is implemented based on the publicly available Caffe platform~\cite{jia:caffe}, which is trained on three NVIDIA GeForce GTX TITAN X GPUs with 12G memory. During training, the learning rate is initialized to 0.01, and during fine-tuning, the learning rate is initialized to 0.001. We train our model using SGD with a batch size of 1 face media set pair, momentum of 0.9, and weight decay of 0.0005.

\subsection{Evaluations on IJB-A Benchmark}
\label{sec4.1}

IJB-A contains 5{,}397 images and 2{,}042 videos from 500 subjects, captured from in-the-wild environment to avoid near frontal bias. For training and testing, 10 random splits are provided by each protocol, respectively. It contains two tasks, face verification and identification. The performance is evaluated by TAR@FAR, FNIR@FPIR and Rank metrics, respectively.

\subsubsection{Ablation Study and Quantitative Comparison}

We first investigate different architectures and losses of MPNet to see their respective roles in unconstrained set-based face recognition. We compare 8 variants of MPNet, \emph{i.e.}, baseline (siamese VGGface~\cite{Parkhi15}), w/o DSG, MPNet$_{\mathrm{K}\in \{3,10,200,500,1000\}}$, and MPNet (backbone: ResNet-101~\cite{he2016deep}).

The performance comparison in terms of TAR@FAR, FNIR@FPIR and Rank1 on IJB-A is reported in the lower panel of Tab.~\ref{tab: Table1}. By comparing the results from w/o DSG \emph{vs.}\ the baseline, around $1\%$ improvement for overall evaluation metrics can be observed. This confirms the benefits of the basic refining tricks in terms of the network structure. Compared with w/o DSG, MPNet$_{\mathrm{K}=500}$ further boosts the performance by around $3\%$, which speaks well for the superiority of using the auxiliary DSG loss to enhance the deep set-based facial representation learning. It simplifies  unconstrained set-based face recognition, yet reserves discriminative and comprehensive information. By varying the numbers of prototypes, one can see that as $K$ increases from 3 to 1{,}000, the performance on the overall metrics improves consistently when $K\leq 500$. This demonstrates that the affinity-based dense subgraph learning of the proposed DSG sub-net can effectively enhance the deep feature capacity of unconstrained set-based face recognition. However, further increasing $K$ does not bring further performance improvement and may even harm the performance on the overall metrics. The reason is that an appropriately large value of $K$ will predict a sparse prototype partition indicator matrix $Z$, which helps reach an optimal trade-off between facial information preserving and computation cost for addressing the large variance and false matching caused by untypical faces. However, an oversize value of $K$ will enforce the learned filters to all zero ones, which always produces invariant performance without any discriminative information. We hence set $K$ to 500 in all the experiments.

For fair comparison with other state-of-the-arts (upper panel of Tab.~\ref{tab: Table1}), we further replace the backbone from VGGface to ResNet-101 (bottom row) while keeping other settings the same. Our MPNet achieves the best results over 10 testing splits on both protocols. This superior performance demonstrates that MPNet is very effective for the unconstrained set-based face recognition in presence of large intra-set variance. Compared with existing set-based face recognition approaches, our MPNet can effectively address the large variance challenge and offer more discriminative and flexible face representations with small computational complexity. Also, superior to the naive average or max pooling of face features, MPNet effectively preserves necessary information through the DSG learning for set-based face recognition.

Moreover, compared with exhaustive matching strategies (\emph{e.g.}, DCNN~\cite{Chen:Unconstrainedfaceverification}) which have $O(mn)$ complexity for similarity computation ($m$, $n$ are media numbers of each face set to recognize) and take $\sim$1.2s for recognizing each probe set, our MPNet is more efficient as it operates on prototype level, which significantly reduces the computational complexity to $O(K^2)$, $K^2 << mn$ ($K$ is the prototype number) and takes $\sim$0.5s for recognizing each probe set. Although naive average or max pooling strategies (\emph{e.g.}, Pooling faces~\cite{Hassner:Poolingfaces}) are slightly advantageous in testing time ($\sim$0.3s for recognizing each probe set), they suffer from information loss severely. Our MPNet effectively preserves the necessary information through DSG learning for unconstrained set-based face recognition. 

\subsubsection{Qualitative Comparison}
\begin{figure}[t]
	\begin{center}	\includegraphics[width=1\linewidth]{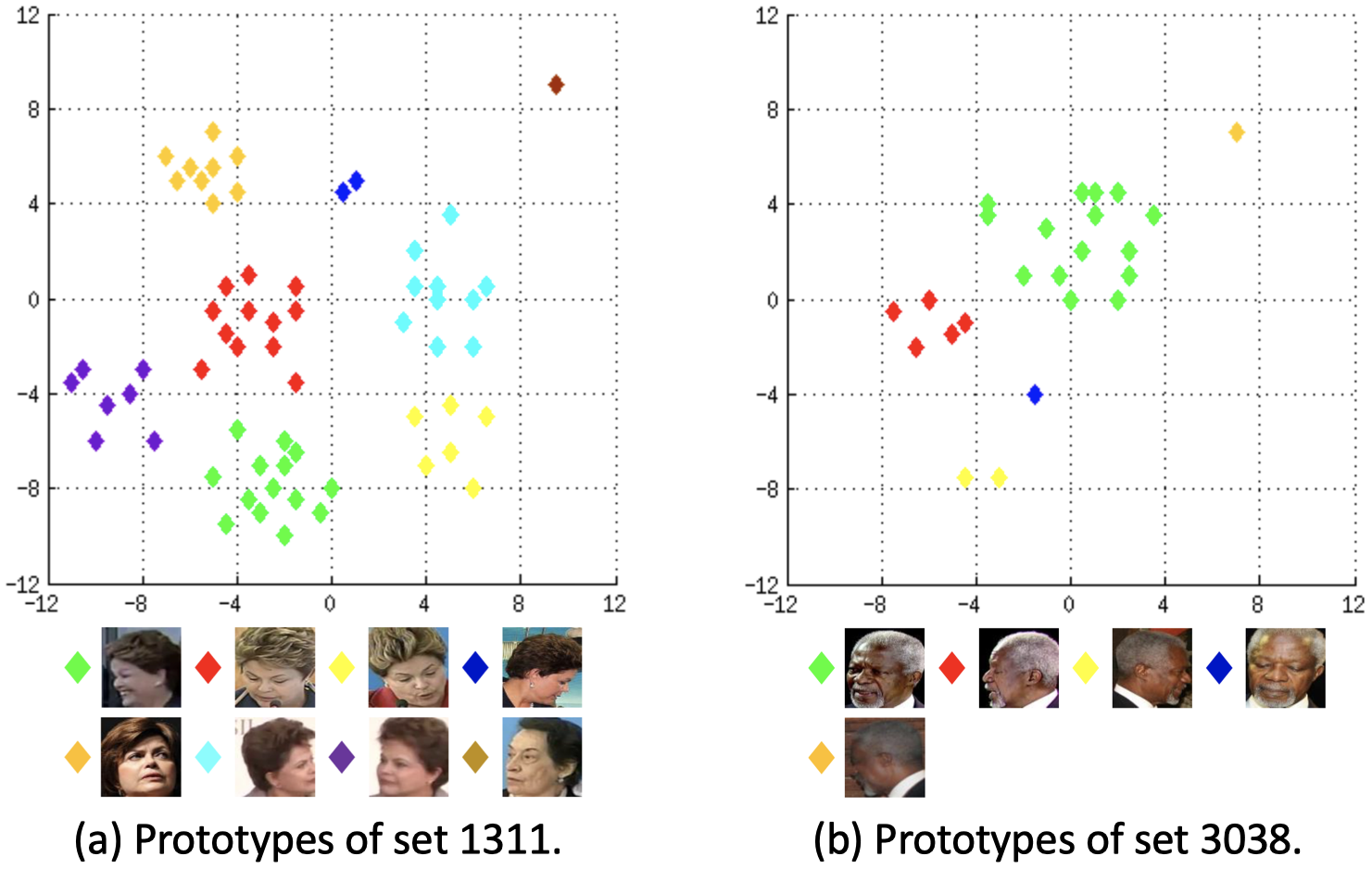}
	\end{center}
	\vspace{-4mm}
	\caption{\small Visualization of learned prototypes within set 1311 (a) and set 3038 (b) by MPNet, from the testing data of IJB-A split1. Each  colored cluster shows a learned prototype. One sampled face within each prototype is shown for better illustration. Best viewed in color.}
	\label{fig: Figure6}
\end{figure}

We then verify the effectiveness of our deep multi-prototype discriminative learning strategy. The predicted prototypes with relatively larger affinities within the set 1311 and set 3038 from the testing data of IJB-A split1 are visualized using t-SNE~\cite{maaten2008visualizing} in Fig.~\ref{fig: Figure6}. We observe that MPNet explicitly learns to automatically predict the prototype memberships within each coarse-level face set reflecting different poses (\emph{e.g.}, the first 6 learned prototypes), expressions (\emph{e.g.}, the $1^{st}$ and the $7^{th}$ learned prototypes), illumination (\emph{e.g.}, the $2^{nd}$ and $5^{th}$ learned prototypes), and media modalities (\emph{e.g.}, the $5^{th}$ and $6^{th}$ learned prototypes). Each learned prototype contains coherent media offering collective facial representation with specific patterns. Outliers within each face set are detected by MPNet (\emph{e.g.}, the last learned prototypes). MPNet is learnt to enhance the compactness of the prototypes as well as their coverage of large variance for a single subject face, through which the heterogeneous attributes within each face media set are sufficiently considered and flexibly untangled. Compared with clustering-based data partition, MPNet with DSG learning is advantageous since it is end-to-end trainable, can learn more discriminative features and is robust to outliers. Learning DSG maximizes the intra-prototype media similarity and inter-prototype difference, resulting in discriminative face representations. This is significantly different from clustering (\emph{e.g.}, k-means) methods where only the similarity defined based on the distance to the center is considered during learning.

\begin{figure*}[t]
	\begin{center} \includegraphics[width=1\linewidth]{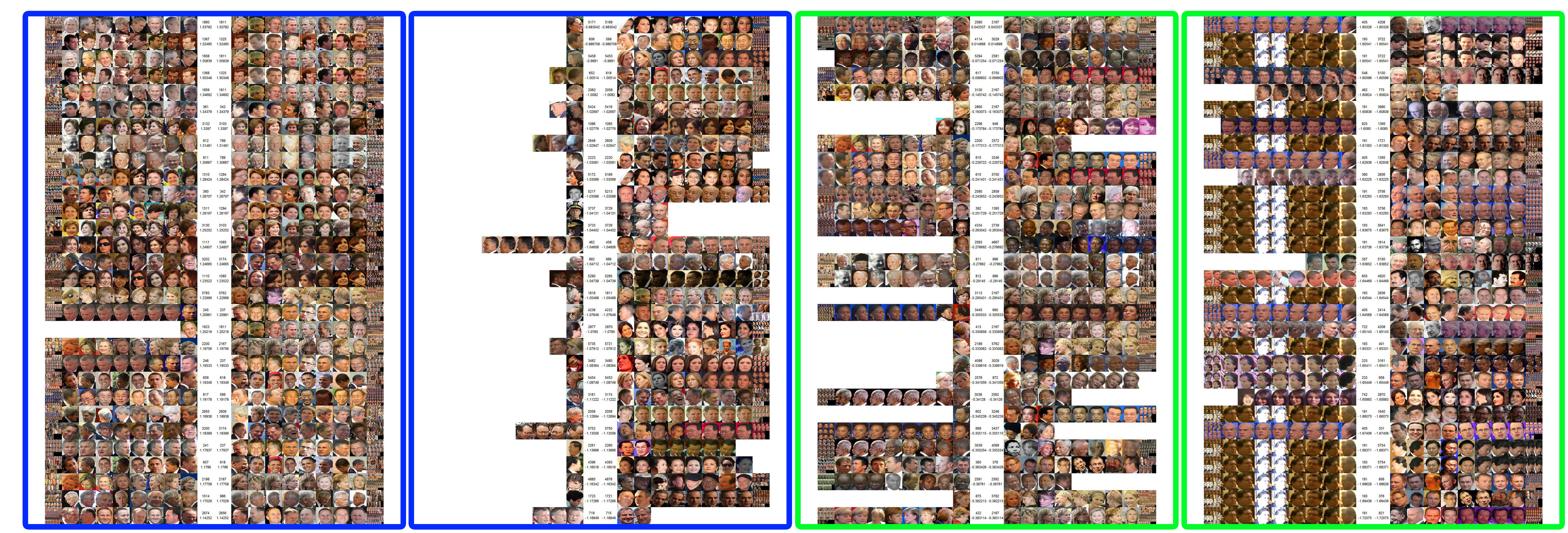}
	\end{center}
	\vspace{-4mm}
	\caption{\small Verification results analysis for IJB-A split1. (blue, left) The best matched cases, (blue, right) The worst matched cases, (green, left) The worst non-matched cases (green, right) The best non-matched cases. For better viewing of this figure, please see the original zoomed-in color pdf file.}
	\label{fig: Figure412}
\end{figure*}

Finally, we visualize the verification results in Fig.~\ref{fig: Figure412} for IJB-A split1 to gain insight into unconstrained set-based face recognition. After computing the similarities for all pairs of probe and reference sets, we sort the resulting list. Each row represents a probe and reference set pair. The original sets within IJB-A contain from one to dozens of media. Up to 8 individual media are shown with the last space showing a mosaic of the remaining media in the set. Between the sets are the set IDs for probe and reference as well as the best matched and best non-matched similarities. Fig.~\ref{fig: Figure412} (blue, left) shows the best matched cases. In the top-30 scoring correct matches, we immediately note that every reference set contains dozens of media. The probe sets either contain dozens of media or one medium that matches well. Fig.~\ref{fig: Figure412} (blue, right) shows the worst matched cases, representing failed matching. The thirty lowest matched results from single-medium probe sets are all under extremely challenging unconstrained conditions. These extremely difficult cases cannot be solved even using the specific operations designed in MPNet. Fig.~\ref{fig: Figure412} (green, left) showing the worst non-matched cases highlights the understandable errors involving single-medium probe sets representing impostors in challenging orientations. Fig.~\ref{fig: Figure412} (green, right) showing the best non-matched cases shows the most certain non-mates, again often involving large sets with enough guidance from the relevant information of the same subject.

\subsection{Evaluations on YTF Benchmark}

YTF contains 3{,}425 videos of 1{,}595 different subjects. The average length of a video clip is 181.3 frames. All the video sequences were downloaded from YouTube. We follow the unrestricted with labeled outside data protocol and report the result on 5{,}000 video pairs.

The face recognition performance comparison of the proposed MPNet with other state-of-the-arts on YTF is reported in Tab~\ref{tab: Table2}. MPNet improves the $2^{nd}$-best by $1.10\%$, which well verified the superiority of MPNet for effectively learning set-level discriminative face representations.

\subsection{Evaluations on IJB-C Benchmark}

\rowcolors{2}{gray!25}{white}
\begin{table*}[t]
	\newcommand{\tabincell}[2]{\begin{tabular}{@{}#1@{}}#2\end{tabular}}
	\tiny
	\begin{center}
		\begin{tabular}{p{3.6cm}<{\centering}|p{2.9cm}<{\centering}|p{2.9cm}<{\centering}|p{2.9cm}<{\centering}|p{2.9cm}<{\centering}}
			\hline
			{\textbf{Method}} & \tabincell{c}{\textbf{TAR@FAR=$10^{-5}$}} & \tabincell{c}{\textbf{TAR@FAR=$10^{-4}$}} & \tabincell{c}{\textbf{TAR@FAR=$10^{-3}$}} & \tabincell{c}{\textbf{TAR@FAR=$10^{-2}$}} \\
			\hline\hline
			GOTS~\cite{maze2018iarpa} & 0.066 & 0.147 & 0.330 & 0.620 \\
			FaceNet~\cite{Schroff:Facenet} & 0.330 & 0.487 & 0.665 & 0.817 \\
			VGGface~\cite{Parkhi15} & 0.437 & 0.598 & 0.748 & 0.871 \\
			VGGface2\_ft~\cite{cao2018vggface2} & 0.768 & 0.862 & 0.927 & 0.967 \\
			MN-vc~\cite{xie2018multicolumn} & 0.771 & 0.862 & 0.927 & 0.968 \\
			\hline
			\textbf{MPNet} & \textbf{0.827} & \textbf{0.898} & \textbf{0.940} & \textbf{0.971} \\
			\hline
		\end{tabular}
	\end{center}
	\vspace{-4mm}
	\caption{\small Face recognition performance comparison on IJB-C.}
	\label{tab: Table3}
\end{table*}

IJB-C contains 31{,}334 images and 11{,}779 videos from 3{,}531 subjects, which are split into 117{,}542 frames, 8.87 images and 3.34 videos per subject, captured from in-the-wild environments to avoid the near frontal bias. For fair comparison, we follow the template-based setting and evaluate models on the standard 1:1 verification protocol in terms of TAR@FAR.

The face recognition performance comparison of the proposed MPNet with other state-of-the-arts on IJB-C is reported in Tab.~\ref{tab: Table3}. MPNet beats the $2^{nd}$-best by 5.60\% w.r.t. TAR@FAR=$10^{-5}$, which further shows its remarkable generalizability for recognizing faces in the wild, and the learned deep features are robust and disambiguated.

\section{Conclusion}

We proposed a novel \textbf{M}ulti-\textbf{P}rototype \textbf{N}etwork (MPNet) with a new \textbf{D}ense \textbf{S}ub\textbf{G}raph (DSG) learning sub-net to address unconstrained set-based face recognition, which adaptively learns compact and discriminative multi-prototype representations. Comprehensive experiments demonstrate the superiority of MPNet over state-of-the-arts. The proposed framework can be easily extended to other generic object recognition tasks by utilizing the area-specific sets. In future, we will explore a pure MPNet architecture where all components are replaced with well designed MPNet layers, which can hierarchically exploit the multi-prototype discriminative information to solve complex computer vision problems.

\section*{Acknowledgement}

The work of Jian Zhao was partially supported by \textbf{C}hina \textbf{S}cholarship \textbf{C}ouncil (CSC) grant 201503170248. 

The work of Junliang Xing was partially supported by the National Science Foundation of Chian 61672519.

The work of Jiashi Feng was partially supported by NUS IDS R-263-000-C67-646, ECRA R-263-000-C87-133 and MOE Tier-II R-263-000-D17-112.

{\small
\bibliographystyle{ieee}
\bibliography{egbib}
}

\end{document}